\documentclass[11pt,3p,review,authoryear]{elsarticle}

\usepackage{amsmath,amsfonts,amssymb}
\usepackage[hidelinks]{hyperref}
\usepackage{booktabs}
\usepackage{xcolor}
\usepackage{color, colortbl}
\definecolor{GrayTop}{HTML}{FFE0B2}
\definecolor{GrayMid}{HTML}{FFF3E0}
\definecolor{LightBlue}{HTML}{549CF1}
\definecolor{AdBlue}{HTML}{000000}

\usepackage{url}

\journal{International Journal of Forecasting}
\biboptions{longnamesfirst}

\begin{document}

\begin{frontmatter}

\title{A probabilistic forecast methodology for volatile electricity prices in the Australian National Electricity Market}

\author[1]{Cameron Cornell}
\ead{cameron.cornell@adelaide.edu.au}
\author[2]{Nam Trong Dinh}
\ead{trongnam.dinh@adelaide.edu.au}
\author[2]{S. Ali Pourmousavi}
\ead{a.pourm@adelaide.edu.au}

\affiliation[1]{organization={The University of Adelaide},
addressline={School of Computer and Mathematical Sciences},
city={Adelaide},
country={Australia}}

\affiliation[2]{organization={The University of Adelaide},
addressline={School of Electrical and Mechanical Engineering},
city={Adelaide},
country={Australia}}

\begin{abstract}

The South Australia region of the Australian National Electricity Market (NEM) displays some of the highest levels of price volatility observed in modern electricity markets. This paper outlines an approach to probabilistic forecasting under these extreme conditions, including spike filtration and several post-processing steps. We propose using quantile regression as an ensemble \textcolor{AdBlue}{tool for} probabilistic forecasting, with our combined forecasts achieving superior results compared to all constituent models. Within our ensemble framework, we demonstrate that \textcolor{AdBlue}{averaging} models with varying training length periods leads to a more adaptive model and increased prediction accuracy. The applicability of the final model is evaluated by comparing our median forecasts with the point forecasts available from the Australian NEM operator, with our model outperforming these NEM forecasts by a significant margin.
\end{abstract}


\end{frontmatter}

\section{Introduction}
Following the deregulation of most of the electricity industries in the 1990s, we saw a growing focus of research on electricity price forecasting (EPF). Decision makers who are properly equipped with reliable electricity price forecasts can adjust their bidding strategies and production/consumption schedule to maximise objectives in day-ahead and real-time trading. Additionally, the electricity market is unique in terms of commodity trade, as it is still economically and technically difficult to store electricity at large scales, thus requiring constant efforts to maintain a balance between production and consumption. The issue of large-scale storage combined with stochastic production and demand behaviour, which depends on time of day, weather conditions, and business activity, leads to extreme spot price volatility, generally unobserved in other commodity markets. As a result, this complex market structure has led to extensive EPF research in recent years \citep{Lago2021forecasting}. 

Recently, there has been a shift in focus from point forecasts (expected value of the spot price) to probabilistic forecasts (estimation of density/interval) \citep{recent}. Probabilistic prediction gained significant momentum after the Global Energy Forecasting Competition (GEFcom2014), which focused solely on probabilistic energy prediction \citep{HONG2016896}. In many applications, such as risk management and bidding in the wholesale electricity market, a successful strategy depends not only on knowledge of expected price levels but also on predicted variability of prices within a given day. The precise method used to capture this variability varies across the literature. Generally, rather than estimating the entire density \citep{conditional2019chai, probabilistic2019afrasiabi}, we limit ourselves to key characteristics of this distribution \citep{Nowotarski2015computing}. A common form of probabilistic forecasting is prediction intervals, where a proportion of the data is expected to lie inside this interval. Alternatively, it can be represented using \(q\) quantiles, where a proportion \(q\) of the data is below this quantile. Both methods can be considered a discretisation of the distribution function to simplify the estimation process. In this study, we focus on quantile estimates in our target range of $0.025 - 0.975$. In particular, we focus on nine key quantiles: $0.025, 0.05, 0.1, 0.25, 0.5, 0.75, 0.9, 0.95$ and $0.975$. 

For a comprehensive review of both point forecasting EPF methods and a more recent review of probabilistic EPF, we refer the readers to the influential papers \citep{WeronLong} and \citep{recent}. In this paper, 
\textcolor{AdBlue}{we propose a model for probabilistic} EPF based on forecast averaging. Many articles have previously concluded that ensemble point forecasting leads to a more accurate and robust prediction of electricity prices \citep{bordignon,NOWOTARSKI2014395,raviv2015forecasting}. However, there is less comprehensive evidence to suggest the benefits of ensemble forecasting in the field of probabilistic forecasting methods. The authors in \citep{Nowotarski2015computing} discussed the possible accuracy gains from \textcolor{AdBlue}{averaging forecasts for probabilistic EPF}, concluding that their ensemble method (dubbed quantile regression averaging (QRA)) leads to more accurate prediction intervals. The results of several forecasting competitions testify to the benefits of forecast averaging, with two of the four winners of the GEFcom 2014 price track using some form of averaging \citep{HONG2016896}, and the number one finding of the 2018 M4 forecast competition is `The improved numerical accuracy of combining' \citep{makridis}. In light of these significant advantages, in this paper, we seek a systematic way of combining sets of increasingly complex price predictions into one final forecast. Our general ensemble method is an extension to the method used by \citep{2019combinething} and QRA, where the contributions are as follows:
\begin{enumerate}
    \item We first show that the probabilistic prediction of the price in the Australian National Electricity Market (NEM) can be improved for many quantiles by filtering the extreme spikes out of the training dataset. Specifically, quantiles 0.10 to 0.95 showed substantial improvements for most models.
    \item Next, we demonstrate two post-processing steps to increase the accuracy of our forecasting methods. Many predictors exhibit ‘oscillatory’ prediction behaviour when trained on highly volatile data. By running a smoothing routine on our predictions, we see a minor increase in accuracy and improved interpretability. Next, we show how autoregression can be used at the probabilistic level to shift the entire estimated density up or down based on the market prices at the start of the 24-hour prediction blocks. The advantage of this additive approach is that it allows us to capture additional temporal dependency when using \textcolor{AdBlue}{models that generate probabilistic predictions}, which are rarely designed in the literature with time series considerations.
    \item During the training stage, we demonstrate that there are systematic accuracy gains by taking our forecast ensembles not only across model types, but also across the length of the training data (memory).
\end{enumerate}

The paper is structured as follows. In section \ref{sec:price_vol_NEM}, we describe our data and the extreme price volatility observed in the NEM and motivate the need for algorithms capable of capturing the complex risk profile observed in South Australia (SA) price series. Section \ref{sec:proposed_methodology} outlines our general methodology, with comprehensive coverage of our constituent models and an outline of the forecast combination framework. Different combination methods are presented in section \ref{section:ensemble}. In section \ref{sec:training_period}, we discuss the combination of models across different training lengths. Finally, the numerical results and the performance comparison from both statistical and economic points of view are presented in sections \ref{sec:Results} and \ref{sec:economic_eva}. The paper is concluded in section \ref{sec:conclusion}.

\section{Data}
\label{sec:price_vol_NEM}
\subsection{The variables}
\label{sec:2.1}
The target variable in this study is the 5-minute interval spot prices in the SA region of the Australian NEM\footnote{Australian NEM publicly available data: \url{https://aemo.com.au/energy-systems/electricity/national-electricity-market-nem/data-nem/market-data-nemweb}}.
The Australian NEM differs from many other countries, such as those depending on the NordPool, as there is no separation into a day-ahead and intraday market. Instead, the Australian NEM operates as a spot market in which prices are determined for each 5-minute trading interval in advance. Generators can rebid anytime until the bids are captured for dispatch for the trading interval \citep{AEMOtimetable}. Prices are then determined by solving a security-constrained linear optimal power flow model \citep{AEMOconsformulation}. Although bidding occurs continuously and can be submitted more than 24 hours in advance, in this study, we restricted our forecast horizon to a maximum of 24 hours for simplicity.

The exogenous variables that we use in our models are primarily historical prices, time, and weather data.

\begin{figure}[t]
\centering
\includegraphics[trim={0cm 0cm 0cm 0cm}, clip, width=0.62\linewidth]{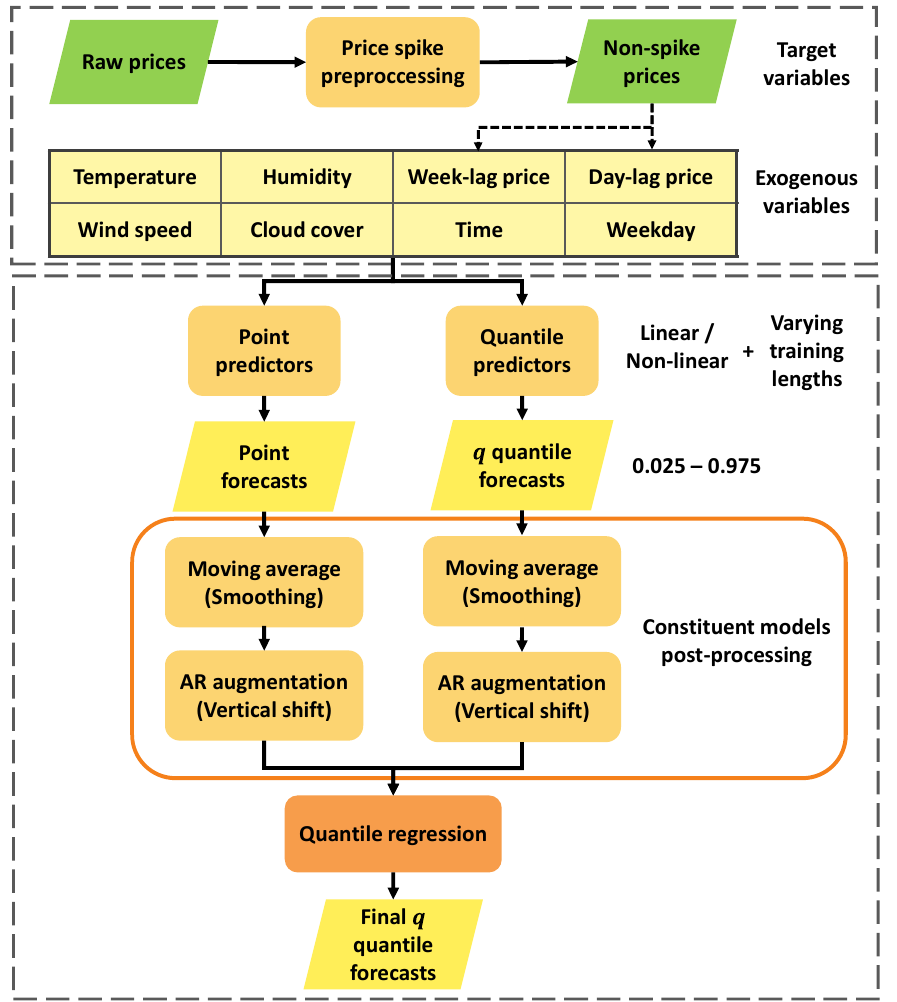}
\caption{Flowchart of the proposed Q-QRA prediction model}
\label{fig:QQRA_flowchart}
\end{figure}

\begin{itemize}
  \item \textbf{Historical electricity prices} consisting of 5-minute spot prices in the SA region (matching our target) are provided by the Australian Energy Market Operator (AEMO). For all given predictor models, we have two lagged prices as model inputs, one from exactly 24 hours ago and another from 7 days ago ($y_{t-288}$ and $y_{t-2016}$).
  \item \textbf{Temporal (time) information} is encoded with an indicator variable for each day of the week, as well as an integer variable between 1 (Time 00:00) and 288 (Time 23:55) corresponding to each of the 5-minute observation times per day.
  \item \textbf{Historical weather forecast data} was collected from World Weather Online \citep{Weather} for three different regions of SA, which are Hallett, Hornsdale and Adelaide. It comprises wind speed (km/h), temperature ($^\circ$C), humidity (\%) and cloud cover (\%) for the testing period.
  \item \textbf{Nonlinear terms} were augmented from the inputs mentioned earlier for the considered linear models. These terms are an order 6 polynomial to the time integer variable to capture the cyclic daily structure, as well as quadratic terms for both temperature and wind speed.
\end{itemize}
The summary of exogenous variables is shown in Figure \ref{fig:QQRA_flowchart}.

\subsection{Price volatility in the NEM}
It has been noted that Australian electricity spot prices exhibit some of the highest volatility levels observed in modern power markets. This is particularly exacerbated within the SA region, where there is a high proportion of renewable energy and reliance on Victorian power production through two interconnectors. This volatility is observed on multiple levels. Not only is the time series generally volatile according to changing weather patterns and price history, but it also exhibits extreme price spikes on an infrequent basis. The intense magnitude of these spikes ensures that while infrequent, they have a significant impact on the market operation and profitability of the market participants. To somewhat limit the effects of these spikes, there is a regulatory ceiling and floor prices, at (AUD) \(-\)\$1000 and \$15,100 per MWh in 2021 \citep{marketpricecap}. Table \ref{voltable} shows price volatility statistics in the SA region over the out-of-sample testing period from 1/1/2018 to 31/12/2021.

\begin{table}[htbp]
    \caption{Summary statistics for the SA electricity spot price from 1/1/2018 to 31/12/2021}
    \centering
    \label{voltable}
    \begin{tabular}{|c|c|c|c|c|c|c|}
    \toprule
    \rowcolor{GrayTop}\textbf{Year} & \textbf{SD}& \textbf{MAD}& \textbf{Mean}& \textbf{Median} &\textbf{Skew} &\textbf{Kurt} \\
    \midrule
    2018 & 346.01 & 20.94 & 92.13 & 82.89 & 33.69 & 1260.03 \\
    \rowcolor{GrayMid} 2019 & 452.23 & 22.76 & 83.04 & 83.50&  25.86 & 777.44 \\
    2020 & 236.04 & 12.67 & 42.93 & 41.75 & 50.10& 2914.50 \\
    \rowcolor{GrayMid} 2021 & 334.30 & 29.55 & 50.70 & 39.18 & 35.65& 1409.57 \\
    \bottomrule
    \multicolumn{7}{l}{$^{\mathrm{a}}$Here MAD is the median absolute deviation, median centred.}
    \end{tabular}
\end{table}

\begin{figure*}[t]
\centering
\includegraphics[trim={0.15cm 1.4cm 0.2cm 1.2cm}, clip, width=\linewidth]{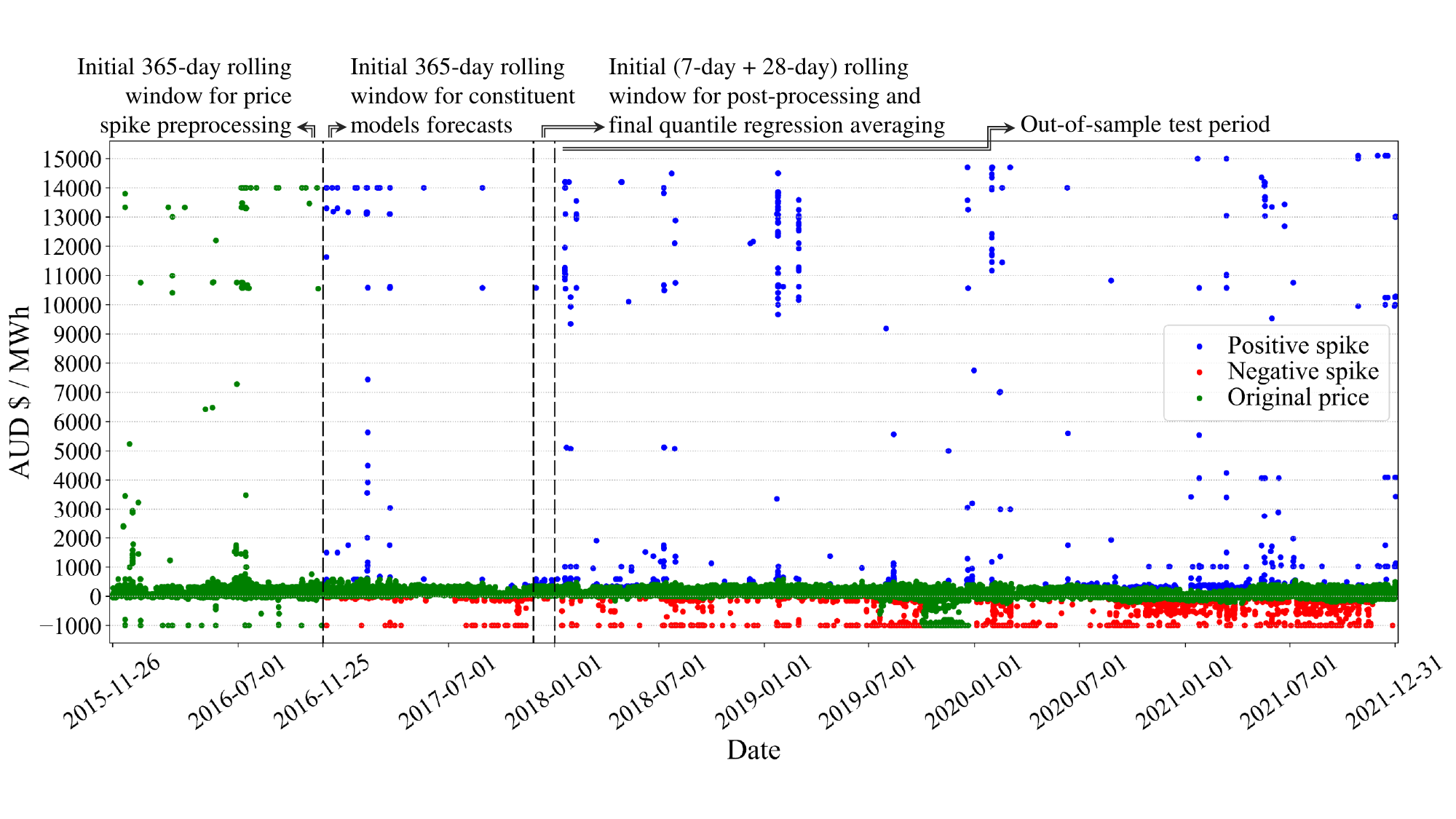}
\caption{The 5-minute interval electricity spot prices in South Australia from 26/11/2015 to 31/12/2021. The vertical dashed lines mark the beginning of the rolling windows, which are identified and labelled within the figure. The colouration of points indicates the spike classifications generated from our preprocessing steps.}
\label{fig:NEM_price_volatility}
\end{figure*}

The numbers presented in Table \ref{voltable} suggest strong evidence of SA price variability. First, we see that the standard deviation (SD) is quite large, being several times higher than the mean and median prices. This contrasts with the median absolute deviations (MAD), which are generally lower than price levels. This behaviour can be explained by the extremely high Kurtosis (fat tails), which ranges around a thousand. We also observe a very heavy skew, indicating that our price distributions are top-heavy. The difference between SD and MAD reflects the empirically observed behaviour: the majority of prices cluster around the distribution centre, with rare extreme outliers. To visualise this behaviour, Figure \ref{fig:NEM_price_volatility} shows the price history over the 4-year testing period, along with the initial rolling training windows and price classifications, as we will discuss in section \ref{sec:proposed_methodology}.

The difference between the implied ‘volatility’ under SD and MAD gives us a glimpse of the primary problem of the data. On the one hand, we see that the SD is a non-robust assessment of deviation from a non-robust centre and reports extremely high variation of the data. On the other hand, MAD seems to be a robust measure of deviation from a robust centre and indicates a quite mild variation in the data. As predictive models are trained by reducing variation statistics, it is critical that we consider the influence of these spikes and the robustness of any variation measure during the training process.

\section{Proposed methodology}
\label{sec:proposed_methodology}

The proposed method is an extension of quantile regression averaging to incorporate not only point forecasts of electricity prices, but also quantile estimates from \textcolor{AdBlue}{multiple probabilistic forecasting models}. Using this method, we can combine the benefits of forecast averaging with a capacity to represent nonlinear relationships at the upper and lower quantile levels of the price density. This is particularly important with our NEM dataset, as there is substantial skew and excess kurtosis (see Figure \ref{fig:NEM_price_volatility}). The goal of our \textcolor{AdBlue}{ensemble model for probabilistic forecasting} is to capture the exogenous conditions that lead to this ‘spiking’ behaviour. In addition to the combination framework, we demonstrate several post-processing methods to improve prediction accuracy. We also show that ensembles can be taken not only across models but also across different subsets of the training data.

The 24-hour prediction process for a given quantile, $q$, is as follows. First, we begin by preprocessing the raw prices by filtering out the extreme price spikes. Using the processed prices as target variables, we then train a series of models to predict the 24-hour ahead electricity prices/desired quantiles based on a series of explanatory variables, such as time of day and weather conditions. Next, we introduce two post-processing steps for these individual model predictions. The first step is to smooth out the prediction series to reduce variability. Then, we augment these predictions by utilising the autoregressive (AR) structure of the residual series to shift the predictions/quantiles based on whether prices are above or below expectation at the most recent observation. As mentioned, all models contain the 24-hour lagged price for each prediction. Therefore, the baseline models are equivalent to a continuous 24-hour ahead rolling prediction. In this context, autoregression can be considered a post-processing step that incorporates the information that is more recent than the previous 24-hour prices used in the base models. In the end, we utilise quantile regression to combine this set of forecasts into our final quantile prediction. The proposed model is, thus, called Q-QRA. Figure \ref{fig:QQRA_flowchart} shows a flowchart of the proposed prediction model. In the following subsections, different aspects of the proposed methodology are explained in detail.

\subsection{Rolling forecast windows}
\label{sec:rolling}
Similar to the works in \citep{DYNAMIC1, DYNAMIC2, DYNAMIC3}, we consider a rolling window scheme in which we retrain our models daily and generate forecasts for the next 24 hours. Although our models are structured to be implemented in some arbitrary 24-hour ahead manner, for testing purposes, we only perform price prediction once a day at midnight. For the first forecast in the out-of-sample test period, i.e., forecast for 1/1/2018, we use data from 26/11/2015 to process and obtain forecasts from the constituent models; see Figure \ref{fig:NEM_price_volatility}. Particularly, the initial 365 days starting from 26/11/2015 are used for preprocessing price spikes. Hence, price classifications are not recorded until 25/11/2016 as shown in Figure \ref{fig:NEM_price_volatility}. The subsequent 365 days are used for generating point/quantile forecasts from the constituent models. Note that since we train the constituent models on different lengths of training data, we visually indicate the size of the rolling window for the constituent forecast models to match the longest training length. We will discuss the varying training lengths in section \ref{sec:training_period}. Finally, the last 35 days prior to the test date are allocated for post-processing and training the final Q-QRA model. To forecast for the next day, that is, 2/1/2018, we shift all the rolling windows forward by one day and repeat the process.

\subsection{Price spike preprocessing}

As discussed in section \ref{sec:price_vol_NEM}, our price series exhibits extremely high levels of volatility and infrequent extreme price spikes. The presence of these spikes ensures the need for robust statistics and consideration of their impact on all levels of model development. As defined in the Introduction, the target quantiles of our prediction routines within this study range from \(0.025-0.975\). As such, we focus on predicting the intra-day variation and mild spiking that occurs with relative frequency. In this context, the extreme spikes (visible in Figure \ref{fig:NEM_price_volatility}) are outliers, and we will show that they can reduce the accuracy of predictions for our quantiles of interest. Therefore, we placed them outside the scope of our predictive goals and looked for spike filtration methods, in which several papers have shown improved accuracy in EPF \citep{JANCZURA201396, CONEJO2005435, WERON2008744}. However, these studies focused on point prediction. Comparatively, we observed that the top 2 winners of the GEFCOM 2014 price track used some form of price filtration for their forecasts \citep{HONG2016896}, indicating that these results extend to probabilistic forecasting.

We now outline our treatment of price spikes, in which the main step is to replace extreme prices with ‘processed’ values. The threshold to consider a given price level as a ‘spike’ can be determined somewhat arbitrarily. In our treatment of spikes, we follow the guidelines found in review articles, e.g., \citep{WeronLong}, while extending these methods to form our classifier. Firstly, defining spikes in terms of a flat threshold leads to inconsistency across long time periods, as general price behaviour is observed to change substantially over time. Secondly, to set varying price thresholds, the measures of price levels should be derived from robust measures, such as median or quantiles. Otherwise, using the rolling average price to define the spikes will be influenced by the spikes. 

Our spike classification method is based on different conditions, each designed to offset what we observed as weaknesses in the other condition. The method combines different quantile levels at different rolling windows, capturing the degree to which the price magnitude is a statistical outlier. Our classification rule for price \(y_t\) at time $t$ is described as follows:
\begin{subequations}
\begin{equation}
    \text{Positive spike}= (y_t>0) \cap (y_t > Q^+_{t,a} + Q^+_{t,m}),
\end{equation}
\begin{equation}
    \text{Negative spike}= (y_t<0) \cap \big( y_t < \frac{Q^-_{t,a} + Q^-_{t,m}}{2} \big).
\end{equation}
\end{subequations}

For our dataset, we have \(Q^+_{t,a}\) and \(Q^-_{t,a}\) as the \(0.975\) and \(0.025\) quantile price of one year of trailing data. Whereas \(Q^+_{t,m}\) and \(Q^-_{t,m}\) represent the \(0.975\) and \(0.025\) quantile price of the month prior to the processing interval. To avoid the impact of long-period negative prices near the end of 2019 on the rolling window values, we take the average of the trailing quantiles for the negative spike filtering.
Figure \ref{fig:NEM_price_volatility} displays the classification from this method in the corresponding colour. Once we have our classification, we replace spikes with the most recent non-spike price, an implementation of the ‘neighbour’ imputation used in \citep{neighbour}.

\subsection{Point forecast}
\label{SVMsection}
Our ensemble method is capable of incorporating quantile forecasts even though it is not limited to them. One of the original benefits of base QRA was to facilitate the use of extensive point prediction research and methods developed in the literature; see, e.g., \citep{WeronLong}. Given that some point prediction methods lack a probabilistic counterpart, we seek to incorporate point predictions into our ensemble model. This allows us to leverage their predictive capacity without having to directly convert the methods into a probabilistic format, which is unlikely to be feasible within the scope of aggregating many forecasts.

In this study, we have selected the radial basis function (RBF) support vector machine (SVM) as the point predictor for our Q-QRA process. Although we have only one point predictor in this paper, our method can accommodate any number of point predictors. Our selection and inclusion of this specific method is largely illustrative and was not targeted as the most accurate point forecasting method.

SVM originates in the classification setting, where they can achieve complex nonlinear boundaries. To create these boundaries, the input data is nonlinearly mapped into higher dimensional kernel space, creating a linear decision boundary in the kernel space. This method was extended by \citep{SVR_NIPS1996_d3890178} to allow regression and has since seen relatively widespread use as a nonlinear regression tool. SVM has been used successfully as the primary point predictor in multiple hybrid EPF models \citep{zhao, CHE20101911}. 

\subsection{Linear quantile regression model}
\textcolor{AdBlue}{Linear quantile regression (QR) has seen widespread application} following its introduction by Koenker and Roger \citep{koenker2001quantile}. It functions as an extension to the least absolute deviation (LAD) regression, where the minimisation of the residual loss function, \(\mathcal{L}_1\), results in an estimate of the conditional median. This extension relies on a further alteration of the loss function from simple residual to the pinball loss function shown below:
\begin{equation}
\label{eqn:pinball_loss}
    \mathcal{L}_{q}(y_t,\hat{y}_{t,q}) = 
\begin{cases} 
      \textcolor{AdBlue}{(y_t - \hat{y}_{t,q}) q} & \hat{y}_{t,q} \leq y_t \\
      \textcolor{AdBlue}{(\hat{y}_{t,q} - y_t) (1-q)} & y_t < \hat{y}_{t,q}. \\
   \end{cases}
\end{equation}

A regression model trained to minimise the expected pinball loss \(\mathcal{L}_{q}\) for a given quantile \(q \in (0, 1)\) can be shown to result in predictions of the conditional quantile \citep{koenker2001quantile}. Predictions are generated by a linear combination of the $m$ predictor variables, as shown below: 
\begin{equation}
    \textcolor{AdBlue}{\hat{y}_{t,q} = \mathbf{X}^\intercal_{t} \pmb{\hat{\beta}}_q},
\end{equation}
where \textcolor{AdBlue}{$\mathbf{X}_{t} = (1,x_{t,1}, x_{t,2}, \dots, x_{t,m})^\intercal$} denotes the input vector and \textcolor{AdBlue}{$\pmb{\hat{\beta}}_{q} = (\hat{\beta}_{q,0}, \hat{\beta}_{q,1}, \hat{\beta}_{q,2}, \dots, \hat{\beta}_{q,m})^\intercal$} denotes the \textcolor{AdBlue}{estimated} coefficient vector of the model, which is determined by the following minimisation process:
\begin{equation}
    \textcolor{AdBlue}{\pmb{\hat{\beta}}_{q} = \text{arg} \min_{\pmb{\beta}_{q}} \sum_{i=1}^l  \mathcal{L}_{q} \big(y_i, \mathbf{X}^\intercal_{i} \pmb{\beta}_{q} \big),}
\end{equation}
\textcolor{AdBlue}{on the $l$ observations in the training data.}
The main limitation of this method is the inability to inherently represent nonlinearities in the data. While we can manually include identifiable nonlinearities, e.g., \(Temperature^2\), and a complex hour structure, it is impossible to capture all such effects exhaustively in this way. This effect is particularly detrimental to our dataset as we observe complex behaviour regarding the outer distributional quantiles. Therefore, it requires the use of  \textcolor{AdBlue}{nonlinear models for generating predictive quantiles} to improve prediction accuracy.

\subsection{Nonlinear quantile regression model}

In order to more accurately capture the risk profile observed in our NEM dataset, we consider \textcolor{AdBlue}{nonlinear models for probabilistic forecasting}. The associated models typically fall under many different labels, such as data mining, machine learning, non-parametric models, etc. 

The nonlinear quantile regression method that we use in this study is the quantile regression forest (QRF) algorithm introduced in \citep{meinshausen2006quantile}. This works in an almost identical manner to the commonly utilised machine learning algorithm of random forest (RF). However, rather than predicting conditional expectations, we estimate the conditional quantile. QRF has recently been applied to forecasting problems, with 3 of 9 final-round contestants from the GEFcom 2017 competition using some forms of quantile forest \citep{hong2019global}. This competition, however, was for load forecasting. To our knowledge, QRF has not yet been used for EPF purposes; hence, we provide an outline of this method below. 

RF and QRF are bootstrap aggregated (bagged) versions of a decision tree. A decision tree is an intuitive model that captures nonlinear relationships between response and predictor variables. To do so, it recursively splits the data into smaller subsets based on binary splits along an explanatory variable. The final product is the original model space separated into multiple distinct rectangular subspaces, each corresponding to a terminal node in the decision tree. To predict the value of a new datapoint, the model allocates the point to the appropriate subspace based on the binary rulings applied to the explanatory variables. Our prediction is the mean of the values in this subspace. We can think of this prediction as a \textcolor{AdBlue}{weighted sum} of the training data \(y_i\).
\begin{equation}
    \textcolor{AdBlue}{\hat{y}_t =\sum_{i=1}^{l} \omega_i(\mathbf{X}_t)y_i,}
\end{equation}
where \textcolor{AdBlue}{\(\omega_i(\mathbf{X}_t)\)} is $1/s$ for the $s$ points in the relevant \textcolor{AdBlue}{terminal} node, 0 elsewhere.

The ‘forest’ aspect of RF is related to the use of bootstrap aggregation to reduce model variance. By training many decision trees based on slightly perturbed data samples, the final prediction can be the average output of each tree. Using the linear combination framework, we have the prediction from our RF as the following:
\begin{equation}
    \textcolor{AdBlue}{\hat{y}_t =\sum_{i=1}^{l} \bar{\omega}_i(\mathbf{X}_t)y_i,}
\end{equation}
\noindent where \textcolor{AdBlue}{\(\bar{\omega}_i(\mathbf{X}_t)\)} is the average of \textcolor{AdBlue}{\(\omega_i(\mathbf{X}_t)\)} across all trees in the forest.

The extension to QRF is \textcolor{AdBlue}{straightforward}. The formal definition of a conditional quantile is \(Q_q(Y_t|\mathbf{X}_t) = \inf(y_t \in \mathbb{R}: q \leq F(y_t|\mathbf{X}_t) ) \). From this we can see that a sufficient value to estimate this quantile is \(F(y_t|\mathbf{X}_t)\) evaluated at finite points. However, \(F(y_t|\mathbf{X}_t)=P(Y_t \leq y_t|\mathbf{X}_t) = \textcolor{AdBlue}{E(\textbf{1}_{(Y_t \leq y_t)}|\mathbf{X}_t)}\). Just as the conditional expectation can be estimated using our weighted combination of \(y_i\), so can this indicator expectation. Using the RF prediction algorithm where \(y_i\) is replaced by the indicator function for \textcolor{AdBlue}{$y_i \leq y_t$}, we arrive at an estimate:
\begin{equation}
    \textcolor{AdBlue}{\hat{F}(y_t|\mathbf{X}_t) =\sum_{i=1}^{l} \bar{\omega}_i(\mathbf{X}_t)\textbf{1}_{(y_i \leq y_t)}}.
\end{equation}
Using this function and the definition of \(Q_q(Y_t|\mathbf{X}_t)\), we can arrive at the prediction $\hat{y}_{t,q}$ for QRF as:
\begin{equation}
\hat{y}_{t,q} = \inf(y_t \in \mathbb{R}: q \leq \hat{F}(y_t|\mathbf{X}_t) ).
\end{equation}
The key benefit of using a QRF over \textcolor{AdBlue}{a linear QR model} is the ability to capture complex behaviour across each quantile. As the QRF model does not estimate any regression coefficient, simply returning values based on empirical quantiles of data points with similar explanatory variable conditions, it has the potential to capture the complex skew risk that we observed in the NEM price series.

\subsection{Constituent models post-processing}

\subsubsection{Prediction smoothing}
The SA price series exhibits high volatility. Consequently, the final predictions are often extremely unstable. This is particularly exacerbated with the outer quantiles of the QRF models, where the model predictions often exhibit ‘oscillatory’ behaviour. To remedy this for application purposes and improve prediction accuracy, we run a smoothing routine over the predicted values. This is particularly helpful before the AR step. Consider a series of consecutive prices closely related such that this period exhibits high autocorrelation. If the predictions during this time are frequently oscillating, we will observe weakly correlated residuals, obscuring the true AR behaviour behind unstable predictions. The technique we use for smoothing is a simple 12\textsuperscript{th} order-centred moving average. Although this may seem extreme given the highly transient nature of prices, we observe that beyond oscillatory behaviour, the prediction volatility is minor within a given hour.

\subsubsection{Density autoregression}

Within our base forecasting routines, we continuously predict prices 24 hours \textcolor{AdBlue}{ahead from the current time $\tau$, i.e., all values $t \in [\tau+1,\tau+288]$} for price series with a 5-minute sampling rate. For each of these forecasts, we have both 24-hour and 7-day lagged prices as input; that is, to forecast noon tomorrow, we are using noon prices today. While this ensures there is always historical price data to perform forecasting in the base models, it fails to utilise the strong AR price behaviour for the earlier predictions within our 24-hour blocks. The predicted values, e.g., \textcolor{AdBlue}{\(\hat{y}_{\tau+1}\), are not using the obviously helpful price \(y_{\tau}\) at time $\tau$}. To capture this key information, we run an AR model on the residual series obtained by subtracting the predicted prices from the true price series. \textcolor{AdBlue}{In a point prediction scenario, this is equivalent to assuming a decomposition of prices at general time} $t$ into:
\begin{equation}
    y_t=f(\mathbf{X}_{t}) + \epsilon_{t},
\end{equation}
\textcolor{AdBlue}{where $f(\mathbf{X}_{t})$ is the output of a function that captures price dependency, and \(\epsilon_t\) is a temporally correlated residual series detailing short-term price trends}. This residual series could be modelled by any uni-variate time series technique, such as ARMA or ARIMA. However, we found equivalent results from simple AR models. \textcolor{AdBlue}{The} final forecast of the price $k$ steps ahead of the current time $\tau$ \textcolor{AdBlue}{would be}:
\begin{equation}
    \hat{y}_{\tau+k}=\hat{f}(\mathbf{X}_{\tau+k}) + \hat{\epsilon}_{\tau+k},
\end{equation}
\textcolor{AdBlue}{where $\hat{f}(\mathbf{X}_{\tau+k})$ is the prediction from a regression model, and $\hat{\epsilon}_{\tau+k}$ represents} the \textcolor{AdBlue}{predicted} residual value from autoregression. Informally, when prices are above/below our baseline forecasts at the beginning of the prediction cycle (24-hour block), we expect prices to remain above/below these levels for some time. All AR models considered are mean (zero) reverting such that \(\hat{\epsilon}_{\tau+k}\xrightarrow[]{}0\) as \(k \xrightarrow[]{} \infty\). This means that our forecast for \(\hat{y}_{\tau+k} \xrightarrow[]{}\hat{f}(\mathbf{X}_{\tau+k}) \) as \(k \xrightarrow[]{} \infty\). The work of \citep{CHE20101911} utilised the AR residual values to increase the robustness of parameter variation in point forecast. However, to the best of our knowledge, this has not yet been applied to probabilistic forecasting, which has multiple forecast ranges. Thus, for the augmentation of our \textcolor{AdBlue}{quantile} predictions, we choose to define this residual series, \textcolor{AdBlue}{$\epsilon_t$}, as the difference between observed prices and the post-smoothing median prediction ($q=0.5$) for that given model. \textcolor{AdBlue}{When making the AR adjusted predictions for $k$ steps ahead of the current time, we  vertically shift the entire set of predicted quantiles from a constituent model by the value $\hat{\epsilon}_{\tau+k}$}. 
\begin{equation}
    \textcolor{AdBlue}{\hat{y}^{\text{p}}_{\tau+k,q} = \hat{y}^\text{s}_{\tau+k,q} + \hat{\epsilon}_{\tau+k},}
\end{equation}
\textcolor{AdBlue}{where $\hat{y}^{\text{p}}_{t,q}$ is the post-processing and $\hat{y}^\text{s}_{t,q}$ is the post-smoothing of $\hat{y}_{t,q}$ from the constituent model. For example}, the vertical shift of all \textcolor{AdBlue}{post-smoothing} quantile predictions in a QRF model is determined by the AR residual prediction of the \textcolor{AdBlue}{post-smoothing} 0.5 quantile prediction of that QRF. \textcolor{AdBlue}{Intuitively, our quantile predictions $\hat{y}^\text{s}_{\tau+k,q}$ describe the density of the prices, with the short-term autoregression simply `re-centering' the quantile predictions with more recent information. As discussed in subsection \ref{sec:rolling}, we elect to make 24-hour ahead forecasts daily at midnight, hence the re-centering occurs daily at midnight.} 

\textcolor{AdBlue}{Overall}, this combination of post-processing approaches allows us to combine a non-inherently time series model (such as our QRF) for long-term, non-recursive forecasting with a simple, inherently temporal model for short-term trends.

\section{Forecast \textcolor{AdBlue}{averaging} with quantile regression}
\label{section:ensemble}
Numerous ensemble methods have been tested within the EPF literature. Among the point forecast methods, simple averaging is a common and effective method. Ordinary least squares (OLS) averaging was introduced in \citep{crane1967two}, with a follow-up paper that generated significant research traction in this area \citep{granger1984improved}. Many papers sought to address the limitations of this method, such as the volatility of beta estimates due to serial correlation of errors and non-robust (\(\mathcal{L}_2\)) fitting routine. The simplest remedy to the second problem was the LAD averaging developed by Nowotarski \citep{NOWOTARSKI2014395}, who eventually developed the quantile regression averaging. The LAD averaging can be seen as the first implementation of QR for averaging, as this is equivalent to a 0.5 quantile regression. Our research follows the direction of these developments and the main ensemble tool in this study is quantile regression. 

\subsection{QRA model}
Quantile regression averaging (QRA) is a probabilistic forecast \textcolor{AdBlue}{combination} method introduced by Nowotarski and Weron in 2015 \citep{Nowotarski2015computing}. It estimates conditional quantiles by applying QR to a set of point forecasts. The improved performance of QRA against its constituent models was verified with several follow-up papers \citep{MACIEJOWSKA2016957, recent, probabilistic2019maciej, UNIEJEWSKI2021105121, conformal2021christopher}. However, its most significant success came from the 2014 GEFcom forecasting competition, where the top two winning teams for the price track used some variant of QRA \citep{macinowgefcom,QRAgef2}.

Mathematically, the model is identical to traditional quantile regression. However, rather than a set of explanatory variables such as temperature and seasonal features, we use the point forecasts \(\hat{y}_{t}\) from multiple predictive models as the explanatory variables. 
\begin{equation}
    \textcolor{AdBlue}{\hat{y}_{t,q} = \mathbf{\hat{y}^{\intercal}}_{t} \pmb{\hat{\beta}}^\text{\tiny QRA}_{q}},
\end{equation}
where \textcolor{AdBlue}{$\mathbf{\hat{y}}_{t} = (\hat{y}^{\text{p}}_{t,1},\ldots, \hat{y}^{\text{p}}_{t,n})^\intercal$} from our $n$ point prediction models \textcolor{AdBlue}{and $\pmb{\hat{\beta}}^\text{\tiny QRA}_{q}$ represents the parameter vector for QRA model}. In QRA, the point predictions are the outputs from the SVM and the 0.5 quantile forecasts (median) of each \textcolor{AdBlue}{quantile regression} model after post-processing, i.e., including the smoothing and autoregression.

\subsection{Q-QRA model}
To arrive at a probabilistic combination tool capable of handling nonlinear volatility features, we extend the model input to include quantile forecasts from the constituent models. Rather than \textcolor{AdBlue}{ combining point forecasts for probabilistic prediction}, this model is a direct combiner of probabilistic forecasts. To do so, the prediction process is mathematically identical to QRA. However, instead of supplying point estimates, we provide estimates of the desired quantile from other probabilistic \textcolor{AdBlue}{forecasting models}:
\begin{equation}
    \textcolor{AdBlue}{\hat{y}_{t,q} = \mathbf{\hat{y}^{\intercal}}_{t,q} \, \pmb{\hat{\beta}}^\text{\tiny Q-Q}_{q},}
\end{equation} 
\textcolor{AdBlue}{where $\pmb{\hat{\beta}}^\text{\tiny Q-Q}_{q}$ represents the parameter vector for Q-QRA model}. This specification deviates from most ensemble algorithms since now each input contains a different feature set for each \(q\)-quantile \textcolor{AdBlue}{$(\mathbf{\hat{y}}_{t}\rightarrow \mathbf{\hat{y}}_{t,q})$}. A review paper on probabilistic forecast averaging tested many quantile forecast combination routines and found that the above specification (labelled QRA-T) achieved the best results \citep{2019combinething}. Our proposed method demonstrates the generalisation capacity of this framework and shows that we can also gain the benefits found in traditional QRA by including point forecasting models. \textcolor{AdBlue}{Hence, $\mathbf{\hat{y}}_{t,q} = (\hat{y}^{\text{p}}_{t,q,1},\ldots,\hat{y}^{\text{p}}_{t,q,p},\hat{y}^{\text{p}}_{t,1},\ldots,\hat{y}^{\text{p}}_{t,n})^\intercal$ consists of $p$ \textcolor{AdBlue}{quantile} prediction models and $n$ point prediction models}. Note that the point forecast inputs of our Q-QRA model only include the output values from the SVM.

Using QR for amalgamating probabilistic forecasts seeks to address the gap between forecast \textcolor{AdBlue}{combination techniques} and the increasingly complex individual methods. We observed that many papers generate sets of predictive quantiles \textcolor{AdBlue}{from nonlinear models}, only to use simple arithmetic averaging to make final predictions. This fails to use the relative performance and correlation of each model in a combining methodology. This issue is addressed in our model. During our testing phase, the QRA and Q-QRA are retrained every day at midnight, with one month of trailing training data.

\section{Selection of training period length}
\label{sec:training_period}

The highly dynamic and non-stationary nature of the NEM electricity prices leads to a myriad of problems for forecasters, one of which is the selection of the appropriate training period lengths. Longer training periods contain a larger volume of information upon which predictions can be obtained. However, in periods where market behaviour departs from historical performance, e.g., due to changes in the market rules, the longer training period will contain mainly inaccurate information for current predictions.

We see this as a realisation of the common variance-bias trade-off. Short-term models display greater reactivity to market change and shifting variable relationships. The cost of this reactivity is a greater tendency to overfit due to the reduced sample size. Longer-term models have a larger sample size and display less overfitting. However, if there are significant changes to the observed market behaviour (such as the consequences of COVID-19), the models will react slowly; thus showing poor performance during this adjustment period. In this sense, short-term models can be considered to be low bias, while longer-term models are low variance. 

With the rapid shift of the energy generation mix to a greater emphasis on renewable energy and the introduction of increasingly complex technologies into the power grid, such as distributed energy resources, we expect many such structural changes to occur in the coming years. Considering these conditions, we suggest EPF practitioners keep in mind this inherent trade-off and consider systematic solutions such as those detailed in this paper. 

Our proposed method for this problem is the combination of models across different training sets. By including models of both short and long training periods in our ensemble, we can leverage the benefits of both stability and reactivity in our forecasts. To do so, in our rolling prediction routines, the QRF models are trained on historical data of one month (Short-QRF), three months of data (Medium-QRF), or one year of data (Long-QRF). The rolling re-training nature of our combining methods ensures that during periods of change, where shorter-term models excel, a larger proportion of the forecast will be attributed to these models (larger co-efficient in the Q-QRA). Conversely, during periods of stability, we expect a higher allocation to the longer-term models. We also expect that different quantiles will have different relationships between accuracy and the length of training data. Under this method, each quantile can use the training data that leads to the most accurate prediction for that quantile. 

Recent research results indicate that there are accuracy gains to be found under these kinds of combination methods, with results showing improved accuracy for simple arithmetic averaging, as well as weighted average schemes based on model performance \citep{DYNAMIC1,DYNAMIC2,DYNAMIC3}. Our implementation relies on inserting these different window models into our already existing model combination scheme, Q-QRA, thus evading additional steps/hyperparameters while retaining the performance of different training length combinations.

\section{Forecasting results}
\label{sec:Results}
\subsection{Evaluation methods}
\label{sec:evaluation_method}

In this paper, we shall follow the probabilistic forecast evaluation criteria discussed in \citep{probabilistic2007gneiting, non2007pinson}. Namely, we seek to maximise sharpness subject to calibration. Sharpness refers to the concentration of the forecast distributions, with sharper distributions providing `narrower' prediction intervals. Calibration, or reliability, ensures that nominal coverage is consistent with empirical coverage. As explained in \citep{gensler2018review}, an important attribute of a successful scoring method is whether it is a ‘proper’ scoring method. Mathematically, this means that the expected loss under this \textcolor{AdBlue}{scoring process is minimised by predictions that are exactly equal to the probabilities/quantiles found in the} true data-generating distribution. The pinball loss function in (\ref{eqn:pinball_loss}) is a proper scoring method \citep{gensler2018review}, and we use it to compare the performance of our predictive models in an out-of-sample manner. Please note that although there are better scoring methods for point predictors, we use the pinball loss function with 0.5 quantile for SVM for consistency with the quantile predictors. The 0.5 quantile pinball loss can also be regarded as half of the mean absolute error.

Another metric that can be used to evaluate the sharpness and reliability of the probabilistic forecast is the continuous ranked probability score (CRPS) \citep{decomposition2000hans,strictly2007tilmann,recent,lauret2019verification}. This score calculates the difference between the cumulative distribution functions (CDF) of the observation and the prediction.
\begin{equation}
\label{eqn:CRPS}
    \text{CRPS} = \int_{-\infty}^{\infty} \big(\hat{F}_t(u) - \textcolor{AdBlue}{\textbf{1}_{(u \geq y_t)}} \big)^2 du,
\end{equation}
where $\hat{F}_t(u)$ denotes the predictive CDF constructed using the nine key quantiles analysed in this study. It is derived based on the \textit{classic} CDF approach described in \citep{lauret2019verification}.

While both the pinball loss function and CRPS can be used to jointly measure the sharpness and reliability of the quantile forecast, it is necessary to consider a more intuitive metric when assessing the quality of the proposed model. To this extent, we consider prediction interval coverage probability (PICP) \citep{lauret2019verification}, which allows us to measure the actual coverage of the prediction interval (PI). Each PI is determined as the interval between the upper and lower quantiles centred around the median. For example, the 95\% PI is defined by the upper 0.975 quantile and the lower 0.025 quantile. Consequently, we want the outcome of the PICP to be as close as possible to the nominal coverage rate. For a $\text{PI}^{(\alpha)}$ with nominal coverage rate $(1 - \alpha )100$\%, the PICP score is given by: 
\begin{equation}
\label{eqn:PICP}
    \text{PICP} = \frac{1}{H} \sum_{t=1}^H c_t,
\end{equation}
where
\begin{equation}
\label{eqn:PICP}
    c_t = \begin{cases} 
      1, & y_t \in \text{PI}^{(\alpha)} \\
      0, & y_t \notin \text{PI}^{(\alpha)},
   \end{cases}
\end{equation}
and $H$ represents the number of test samples.

In the following subsections, we first investigate the effects of the proposed spike treatment on the constituent predictive models by comparing the pinball loss evaluations of the raw and processed series. Then, we show that the prediction can be improved by applying smoothing and AR augmentation on the outputs of each model. Also, we evaluate the effect of the dynamic training length at different periods over the considered horizon. Lastly, we show the performances of different ensemble routines and make a comparison with the available forecast model from AEMO.

\subsection{Spike-processing}

\begin{figure}[t]
\centering
\includegraphics[trim={0cm 0cm 0cm 0cm}, clip, width=0.85\linewidth]{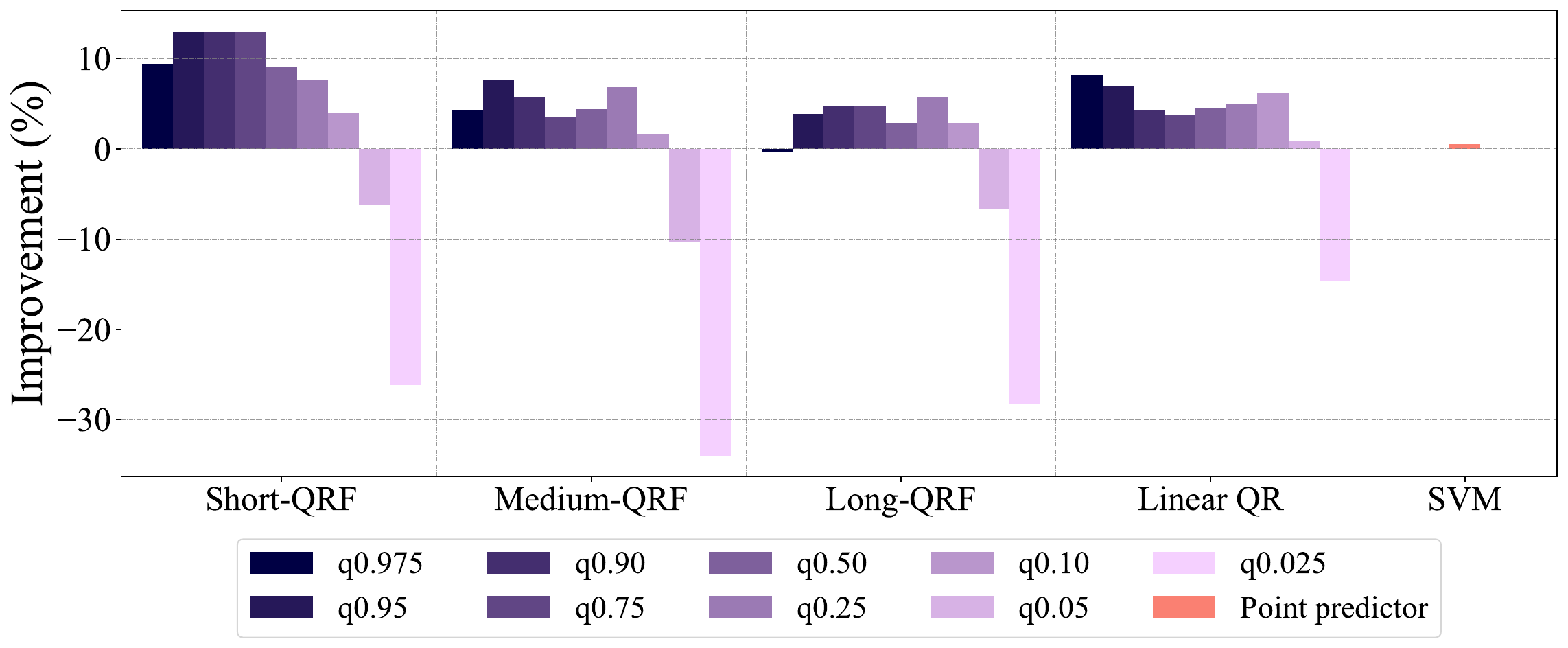}
\caption{The improvements in pinball losses (\% reduction) of constituent models after removing spikes using the pre-processing steps. The colour gradient indicates the different quantiles, read left to right in descending quantile order.}
\label{fig:raw_vs_processed}
\end{figure}

To demonstrate the harmful effects of extreme price spikes and investigate the benefits of our spike processing routines, we train all models twice, once on the raw data and then on the processed price series. Figure \ref{fig:raw_vs_processed} shows the relative improvements when training on the processed prices compared with the raw counterparts. The results demonstrate that even when using inherently robust methods such as linear quantile regression, the presence of extreme spikes in the training data reduces the accuracy of predictions within most quantiles. An observed exception to this is the extremely lower quantiles, where the raw models achieved a higher degree of accuracy (negative improvement values). This is due to the frequent occurrences of negative spikes in the later years; thus entering our target quantiles. 
Overall, these results indicate a moderate but consistent benefit to extreme spike filtration. For those desiring a wider target quantile range, it is perhaps best to use the filtered series for the inner quantiles and reserve the original, spike-inclusive series for the extreme ranges. However, for consistency, the remaining training routines are performed using the filtered series.

\subsection{Smoothing and autoregression}

\begin{figure}[t]
\centering
\includegraphics[trim={0cm 0cm 0cm 0cm}, clip, width=0.85\linewidth]{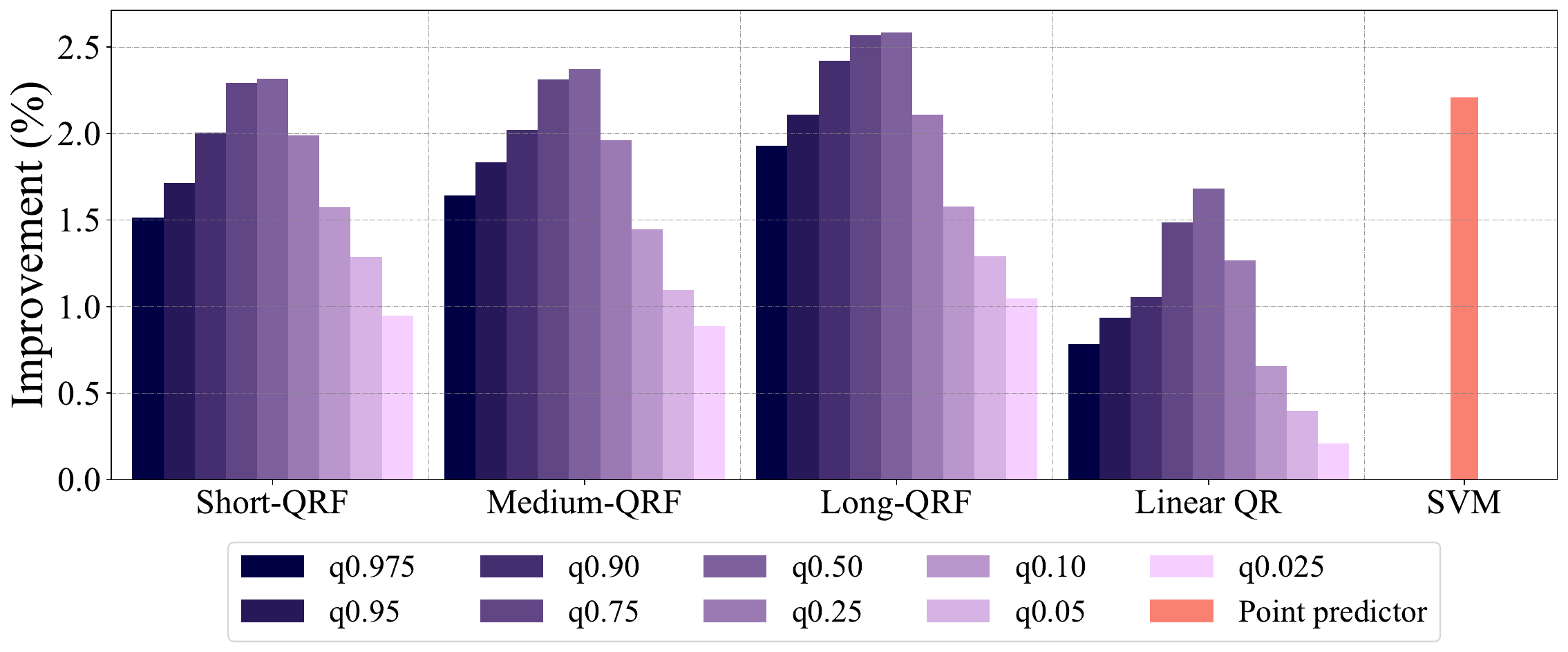}
\caption{The improvements in pinball losses (\% reduction) of constituent models after the post-processing steps of AR adjustment and forecast smoothing. The colour gradient indicates the different quantiles, read left to right in descending quantile order.}
\label{fig:post_vs_processed}
\end{figure}

Upon obtaining the outputs of the constituent models, the prediction values are smoothed out from a 1-hour window moving average and vertically shifted using AR augmentation. The effects of this post-processing step can be seen in Figure \ref{fig:post_vs_processed}, which shows minor improvements for all quantile levels, generally around 1--2.5\%. 

\subsection{Constituent models}
\begin{figure}[!h]
\centering
\includegraphics[trim={0cm 0cm 0cm 0cm}, clip, width=0.74\linewidth]{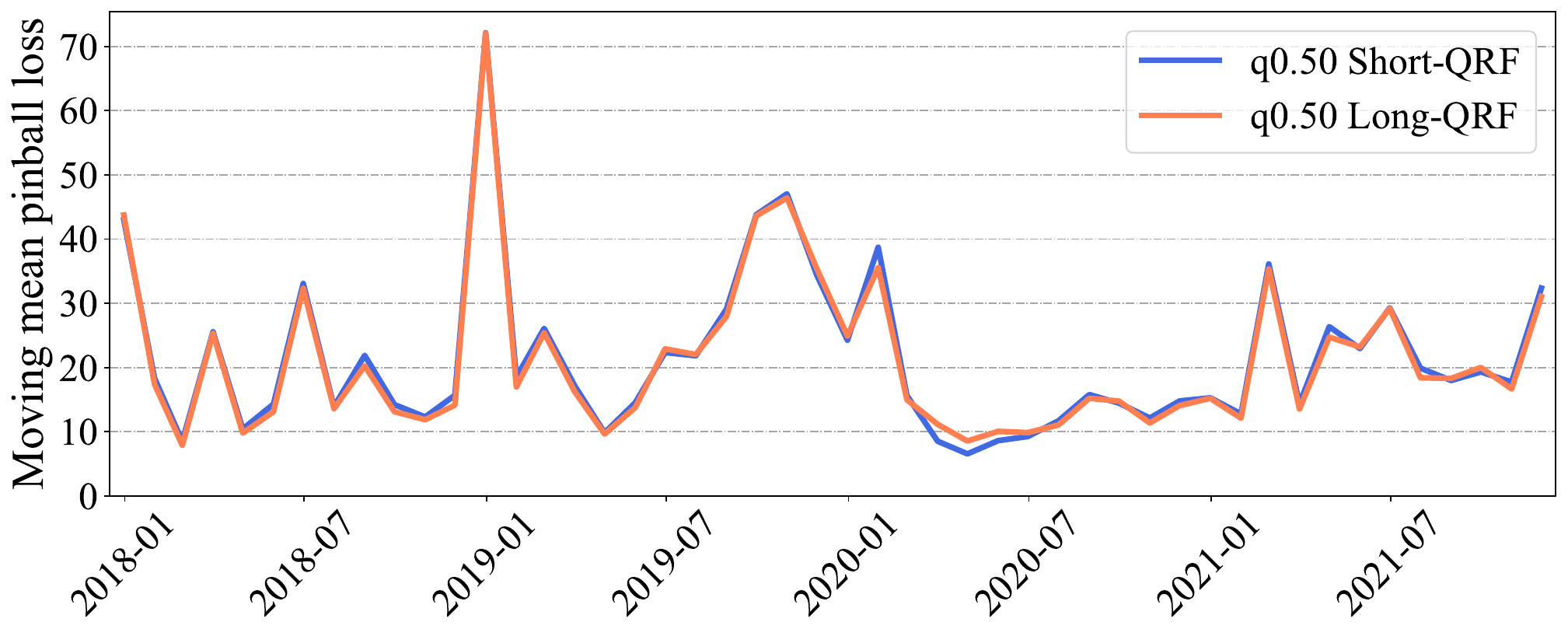}
\caption{The pinball loss 28-day moving average for 0.5 quantile (median) forecasts by the Short-QRF and Long-QRF models, spanning from 1/1/2018 to 31/12/2021.}
\label{fig:moving_pinball}
\end{figure}

Table \ref{tab:pinball_constituent} shows the average pinball losses of each constituent model at different quantile levels from the four-year testing period. At first glance, the Short-QRF model, which is trained on one month of rolling data, appears to have the worst prediction accuracy. Especially when compared to the Long-QRF model, with the one-year rolling model outperforming in all tested quantiles. However, Figure \ref{fig:moving_pinball} shows that the superiority of the Long-QRF model is simply a result of a lower average value across the entire assessment period. It can be seen that the short-memory model systematically beat the longer-term model at around the beginning of 2020, which is the start of COVID-19. Then later that year, once sufficient data was gathered to ‘learn’ about the impact of pandemic, the long-memory model performed slightly more accurately than the shorter model. Therefore, despite common sentiment in non-time-series based regression, more data does not necessarily imply more accuracy.

These findings suggest that the benefit of combining models with different training lengths may not simply be the variance reduction seen in traditional ensemble methods but an increased capacity for adaptation when significant changes occur in the market. An ensemble routine would ideally leverage longer memory models during periods of stability and shift model weights towards shorter training periods under periods of market changes.

\begin{table}[!t]
    \caption{The mean pinball loss associated with the probabilistic forecasts of each constituent model. Boldface values indicate the top performing model for a specific quantile}
    \centering
    \begin{tabular}{c|cccc}
    \toprule
    \rowcolor{GrayTop}\textbf{Quantile} &
    \textbf{Linear QR}&
    \textbf{Short-QRF}& \textbf{Medium-QRF}& \textbf{Long-QRF} \\
    \midrule
    0.025 & \textbf{9.87} & 10.19 & 10.06 & 10.17 \\
    \rowcolor{GrayMid} 0.05 & \textbf{11.36} &  11.71 & 11.56 & 11.63 \\
    0.10 & \textbf{13.64} & 13.92 & 13.71 &  13.80 \\
    \rowcolor{GrayMid} 0.25 & 17.96 &  18.25 & \textbf{17.90} & 18.01 \\
    0.50 & \textbf{20.99} & 21.55 & 21.06 & 21.18 \\
    \rowcolor{GrayMid} 0.75 & \textbf{20.31} & 20.78 & 20.34 &  20.41 \\
    0.90 & 17.11 & 17.15 & 16.91 & \textbf{16.88} \\
    \rowcolor{GrayMid} 0.95 & 14.85 & 14.74 & 14.54 & \textbf{14.49} \\
    0.975 & 12.98 & 12.94 & 12.73 & \textbf{12.67} \\
    \specialrule{0.05em}{0em}{0em}
    \rowcolor{GrayMid} \textbf{SVM} & \multicolumn{4}{c}{22.94} \\
    \bottomrule
    \end{tabular}
    \label{tab:pinball_constituent}
\end{table}

In addition, among the constituent models we see in Table \ref{tab:pinball_constituent}, the nonlinear \textcolor{AdBlue}{QR} models (QRF) achieve the best performance in the upper outer quantiles (0.9, 0.95 and 0.975 quantiles) where linear QR struggles. Whereas, by a small margin, linear QR achieves the highest scores at the 0.50 and 0.75 quantiles. Although the low quantiles display results favouring the linear QR, these are somewhat skewed due to the frequent occurrences of negative spikes, which are not accurately captured due to the process of spike filtration. Overall, the relative performance of these models supports our hypothesis that much of the nonlinearities occur specifically in the outer distributional quantiles. 

\subsection{Different ensemble routines}

\begin{table}[t]
    \caption{The mean pinball loss associated with the probabilistic forecasts of each averaging routine, together with the score for the `Best constituent', representing the highest performing individual model for each quantile (the boldface values from Table \ref{tab:pinball_constituent}). Boldface values indicate the top-performing model for a specific quantile.}
    \centering
    \begin{tabular}{c|ccc}
    \toprule
    \rowcolor{GrayTop}\textbf{Quantile} &
    \textbf{Best constituent}&
    \textbf{QRA}& \textbf{Q-QRA} \\
    \midrule
    0.025 & 9.87 & 10.0 & \textbf{9.59} \\
    \rowcolor{GrayMid} 0.05 & 11.36 & 11.52 & \textbf{11.14} \\
    0.10 & 13.64 & 13.72 & \textbf{13.29} \\
    \rowcolor{GrayMid} 0.25 & 17.90 & 17.69 & \textbf{17.45} \\
    0.50 & 20.99 & \textbf{20.42} & \textbf{20.42} \\
    \rowcolor{GrayMid} 0.75 & 20.31 & 19.76 & \textbf{19.68} \\
    0.90 & 16.88 & 16.82 & \textbf{16.49} \\
    \rowcolor{GrayMid} 0.95 & 14.49 & 14.63 & \textbf{14.25} \\
    0.975 & 12.67 & 12.86 & \textbf{12.48} \\
    \bottomrule
    \end{tabular}
    \label{tab:pinball_ensemble}
\end{table}

To investigate the possible gains from averaging over the models, we compare \textcolor{AdBlue}{the probabilistic forecasts} from two ensemble models (QRA \& Q-QRA) against the best performing constituent model for each quantile. Table \ref{tab:pinball_ensemble} displays the average pinball loss for these ensembles and constituents. Since the QRA routine only leverages the median forecasts from linear and QRF models as well as point predictions from constituent models, the improvements are generally concentrated at the inner quantiles and fail to capture the nonlinearities at the outer quantiles; thus, resulting in worse predictions. Whereas for the Q-QRA model, the results are commensurate with much of the literature surrounding the benefits of forecast \textcolor{AdBlue}{averaging}. The proposed Q-QRA model attains the minimum pinball losses at all investigated quantiles. A similar pattern is observed in Table \ref{tab:CRPS} for the CRPS metric, with the Q-QRA model again achieving the lowest losses.

\begin{table}[t]
    \caption{The continuous ranked probability score (CRPS) of the probabilistic forecasts for the individual models together with the averaging routines. The boldface value indicates the top-performing model.}
    \centering
    \begin{tabular}{cccccc}
    \toprule
    \rowcolor{GrayTop} \textbf{Short-QRF} &
    \textbf{Medium-QRF} & \textbf{Long-QRF} & \textbf{Linear QR} & \textbf{QRA} & \textbf{Q-QRA}\\
    \midrule
    39.26 & 38.91 & 38.51 & 38.14 & 37.17 & \textbf{37.02} \\
    \bottomrule
    \end{tabular}
    \label{tab:CRPS}
\end{table}

To validate these observations, we utilised the Diebold and Mariano test \citep{diebold} to evaluate the statistical significance of the disparity in pinball loss observed in Table \ref{tab:pinball_ensemble} between the Q-QRA and QRA methods. It was found that for all non-median quantiles assessed in Table \ref{tab:pinball_ensemble}, the difference in forecast performance was statistically significant at a 5\% level of significance in a one-sided test (the median forecasts are identical across these models by construction). Beyond this aggregated test that assessed all dispatch periods together, we also ran the test on each dispatch period individually. Table \ref{tab:performance_diff_across_periods} records the number of daily dispatch periods (out of 288) where the Q-QRA model had statistically significantly lower pinball loss. There is notable variation in the number of passed periods across the relevant quantiles. However, even with this variation, the minimum number of periods passed greatly exceeds what would be expected if the QRA and Q-QRA models had identical forecast performance. Under the null hypothesis of equal forecast accuracy, the Q-QRA is expected to pass between 7 and 22 periods with 95\% confidence. This range is derived from $np\pm 1.96\sqrt{np(1-p)}$, which is determined using a normal approximation for the binomial result (either passing or not passing within each of the 288 dispatch periods). The lower half of the quantiles had a greater number of periods passing the test, with the upper half of the quantiles showing a larger performance difference for the outer quantiles. These results are consistent with Table \ref{tab:pinball_ensemble}, with the number of passed periods generally relating to the size of the difference in the pinball score.

\begin{table}[t]
    \caption{The number of daily dispatch periods (out of 288) where the Diebold and Mariano test reveals statistically significant discrepancies in the forecasting accuracy between the QRA and Q-QRA models at 5\% significance level}
    \centering
    \begin{tabular}{c|c c c c c c c c c}
    \toprule
    \rowcolor{GrayTop} \textbf{Quantile} &
    \textbf{0.025} & \textbf{0.05}& \textbf{0.1} & \textbf{0.25}& \textbf{0.5}& \textbf{0.75} & \textbf{0.9}& \textbf{0.95}& \textbf{0.975} \\
    \midrule
    \# \textbf{Significant} & 218 & 228 & 269 & 204 & - & 57 & 91 & 160 & 194\\
    \bottomrule
    \end{tabular}
    \label{tab:performance_diff_across_periods}
\end{table}

\begin{table}[!h]
    \caption{The prediction interval coverage probability (PICP) of the probabilistic forecasts for the individual models together with the averaging routines. Boldface values indicate the top-performing model for a specific interval.}
    \centering
    \begin{tabular}{c|cccccc}
    \toprule
    \rowcolor{GrayTop} \textbf{PI} & \textbf{Linear QR} & \textbf{Short-QRF} & \textbf{Medium-QRF} & \textbf{Long-QRF} & \textbf{QRA} & \textbf{Q-QRA}\\
    \midrule
    50\% & 0.469 & 0.420 & 0.454 & 0.460 & \textbf{0.470} & \textbf{0.470} \\
    \rowcolor{GrayMid} 80\% & 0.746 & 0.693 & 0.732 & 0.741 & 0.756 & \textbf{0.759} \\
    90\% & 0.841 & 0.800 & 0.831 & 0.841 & 0.854 & \textbf{0.858} \\
    \rowcolor{GrayMid} 95\% & 0.891 & 0.862 & 0.886 & 0.895 & 0.905 & \textbf{0.908} \\
    \bottomrule
    \end{tabular}
    \label{tab:PICP}
\end{table}

\begin{figure*}[!t]
\centering
\includegraphics[width=0.98\linewidth]{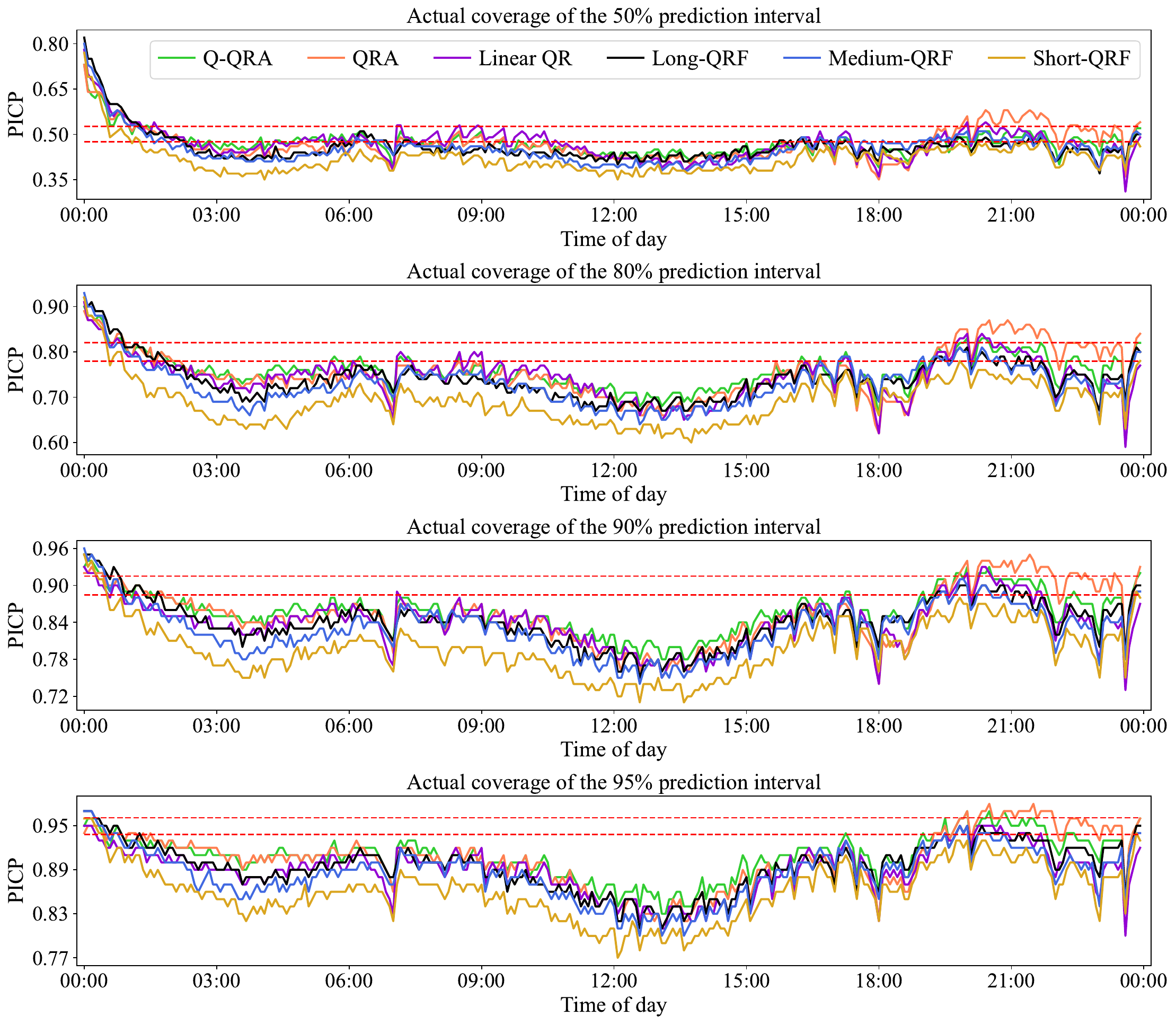}
\caption{The prediction interval coverage probability (PICP) for all models across the 288 dispatch periods in the 24-hour cycle. The horizontal red dashed lines serve as visual aids, indicating consistency bars associated with a 95\% confidence level.}
\label{fig:PICP_time}
\end{figure*}

In addition to pinball losses, we also compare the coverage performance of each ensemble routine. The relative performance, indicated by the PICP shown in Table \ref{tab:PICP} is consistent with the pinball loss results. Overall, the results show that empirical coverage was below target for all models and prediction intervals. Nonetheless, our proposed Q-QRA model consistently achieved the smallest gap between the PICP score and the nominal coverage. Figure \ref{fig:PICP_time} displays the PICP for these intervals throughout the day to see if there is seasonality in the reliability metrics. To provide intuition around the scale of these trends, the plots include consistency bars with a 95\% confidence level \citep{increasing2007brocker}. These bars indicate the expected variation of the PICP from the nominal coverage due to the inherent randomness caused by finite test data. Given that the observations within each specific 5-minute interval are spaced 24 hours apart, we consider low serial correlation for these intervals. Consequently, we calculate the consistency bars using the binomial distribution, which is identified as the null-distribution for non-serially correlated coverage indicators \citep{cons_bars}. Note that these lines only serve as visual aids; a formal evaluation of the coverage is provided through the Kupiec results detailed below. We see that reliability is excessively high at the start of the 24-hour cycle as the AR boosting `re-centers' the density, which likely improves coverage. The quantiles are not trained conditional on the AR improvements; hence they are unaware of the excess coverage during these periods and could have shrunk the quantiles inward to reflect the increased accuracy of this short-term prediction augmentation (the post-processing solely affects reliability without altering sharpness). During the rest of the dispatch period, the PICP is consistently below target, roughly matching our previous PICP results that were aggregated across time periods. To formally analyse reliability across the dispatch periods, we conduct a two-sided hypothesis test to assess whether empirical coverage is equal to nominal coverage. Specifically, we employ the Kupiec coverage test from \citep{kupiec} at the 5\% level of significance for each quantile (0.025, 0.05, 0.1, 0.25, 0.5, 0.75, 0.9, 0.95, 0.975) and for every dispatch period across all models.
For the Q-QRA model, the test was passed in 15--47\% (27.8\% on average) of the dispatch periods across the nine quantiles. For the QRA model, the test was passed in 15--44\% (26.7\% on average) of the cases, which approximately coincides with the 15--20\% cases (across 24h and 90\% or 50\% PI) reported in Table 1 in \citep{UNIEJEWSKI2021105121} for the Polish and Scandinavian markets. The forecasts for the constituent models were generally lower, particularly for the Short-QRF, which achieves the lowest score. In general, the test was passed more times for higher quantiles than for lower ones. Overall, all forms of reliability analysis suggest systematic under-coverage for all model types, however, the level of this under-coverage approximately matches other instances of EPF coverage analysis. This suggests that coverage remains a unique challenge, and stands as a strong candidate for future research. In terms of the relative performance of reliability metrics, the Q-QRA can be seen to generally achieve the highest reliability for most of the prediction cycles and intervals.

\subsection{AEMO pre-dispatch price comparison}
\label{sec:AEMO_price_forecast}

\begin{figure*}[t]
\centering
\includegraphics[trim={0cm 0cm 0cm 0cm}, clip, width=0.9\linewidth]{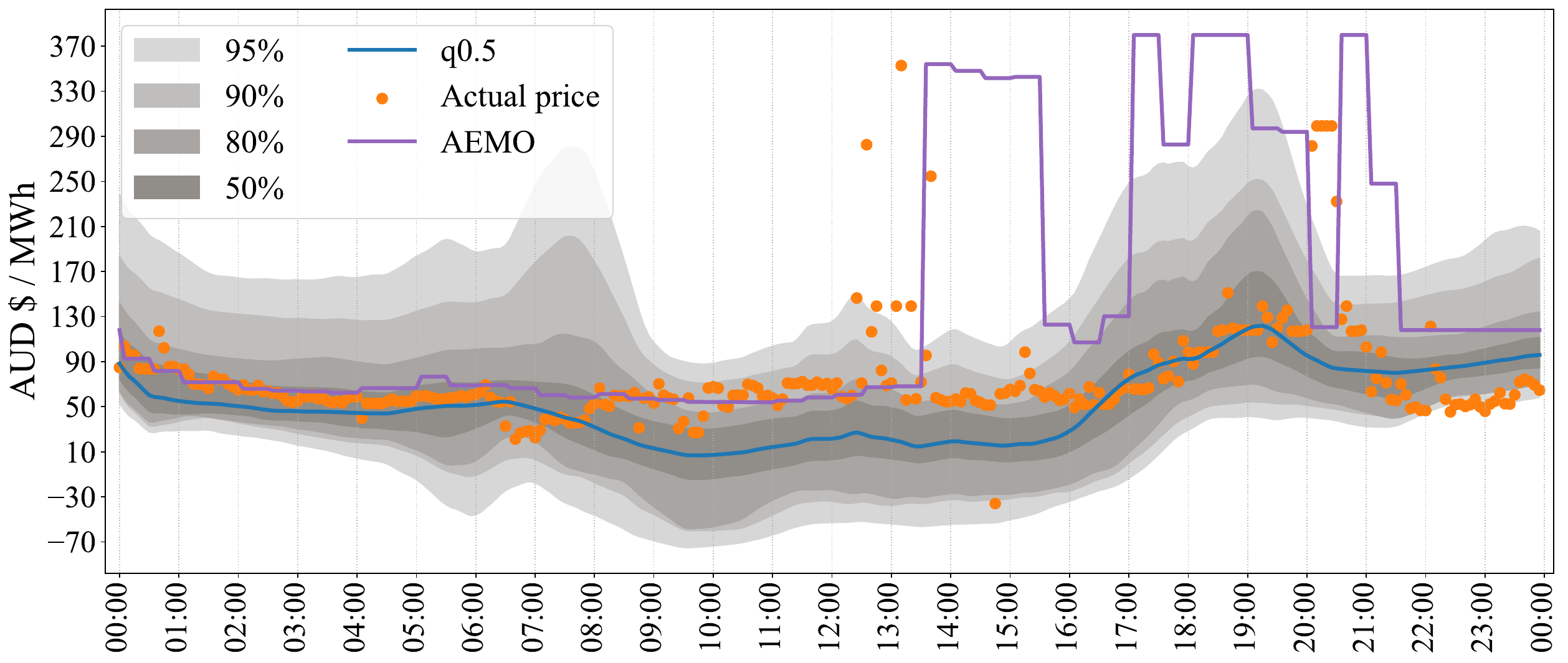}
\caption{The electricity price forecasts in the South Australia region for December 1, 2021. The AEMO's predicted values are presented as a solid purple line, while the Q-QRA model's median forecast is marked with a blue line, and the predictive density is illustrated through various degrees of shading.}
\label{fig:quantile_crossing}
\end{figure*}

In addition to providing the spot prices from the historical dispatch intervals, AEMO must prepare and publish a pre-dispatch schedule containing 30-minute pre-dispatch (forecast) wholesale prices until the end of the next trading day. According to \citep{AEMOPreDispatch}, these pre-dispatch prices represent the expected bids at the last trading interval in each 30-minute period, i.e., pre-dispatch prices are only available at 00:00, 00:30, etc. Although these 24-hour ahead forecasts are updated every half hour, we select and evaluate the day-ahead forecasts made at 00:00 every day to align with the prediction horizon of our Q-QRA implementation. Note that our proposed model can perform a rolling forecast every 5 minutes. However, as mentioned earlier, our results are based on daily predictions so that all 24 hours of predictions are included in the evaluation metrics. Due to the availability of prediction data from AEMO, only twelve months of data from the beginning of 2021 until the end of the year is used. 

\begin{table}[t]
    \caption{The mean and median absolute prediction error for the AEMO pre-dispacth forecast and the 0.5 quantile (median) forecast from the proposed Q-QRA model in 2021.}
    \centering
    \begin{tabular}{c|cc}
    \toprule
    \rowcolor{GrayTop}\textbf{Predictors} &
    \textbf{Mean AE} &
    \textbf{Median AE}\\
    \midrule
    AEMO pre-dispatch & 223.82 & 18.17 \\
    \rowcolor{GrayMid} Q-QRA median & \textbf{41.12} & \textbf{17.26} \\
    \bottomrule
    \end{tabular}
    \label{tab:AEMO_prediction}
\end{table}

Table \ref{tab:AEMO_prediction} shows the comparative performance between the point forecast from AEMO and the median forecast of the proposed model in terms of the mean absolute error (AE) and median AE (in \$AUD/MWh). The results show that our median forecast outperforms AEMO in the half-hourly prediction period. The high mean AE of the prediction of AEMO can be explained in Figure \ref{fig:quantile_crossing}, which displays the probabilistic electricity price forecasts for the next 24 hours on the first day of December 2021. It can be seen that although the AEMO model (purple line) thrives in the first few hours, its predictions become significantly unstable in the second half of the forecast horizon. As a result, the AEMO model has a mean AE 12 times higher than its median AE and 5 times higher than the mean AE of our point forecast (blue line).

\section{Economic evaluation}
\label{sec:economic_eva}

Although achieving a lower prediction error is an important goal in price prediction, the true impact of the accuracy of the forecast must be evaluated in decision-making processes \citep{energy2020hong}. For example, in \citep{electricity2018Chitsaz}, the authors developed a mixed-integer linear programming (MILP) problem for battery operation using price forecasts. Specifically, the model was developed to maximise the savings for a microgrid through energy arbitrage of a central battery system. In our model, however, instead of considering a central battery, we assume that each household has a home battery system, e.g., Tesla Powerwall \citep{powerwall}. This is to ensure that we consider variability in different energy profiles.

In Australia, there are several newly emerged electricity retailers, e.g., Amber Electric \citep{amber}, allowing residential prosumers to buy and sell electricity at spot prices. Therefore, in our model, we consider residential households to be the decision makers that operate their battery system to minimise their electricity bill. To provide a simple case study, we generate price forecasts once a day at midnight for one day ahead, making it easier for the prosumers to manage and monitor their home battery system. Since price forecasts are updated at midnight, the optimisation model only needs to be solved once a day to find the optimal scheduling of the home battery for the following day.

\subsection{Prosumers' electricity cost minimisation problem}
We assume that each household has a rooftop solar photovoltaic (PV) system, and the generated electricity can be used to satisfy household consumption, stored in the home battery or sold back to the grid. Using the home battery, electricity prosumers can exercise energy arbitrage not only to reduce their electricity bill, but also to make a profit by selling electricity back to the grid during high price intervals. The optimisation model for each prosumer for each day is as follows:
\begin{subequations}
\label{eqn:CBS_operation}
\begin{flalign}
\label{eqn:obj_func}
     \min_{\theta}& \; \text{Cost} = \sum_{t \in T} \delta_t (n^+_t - n^-_t) + \lambda_t n^+_t \\
    \text{s.t}. \notag \\
\label{eqn:battery_soc}
    \quad \quad & E_t = E^{\text{init}} + \sum_{h=1}^t \big( P^{\text{ch}}_h - \frac{1}{\Gamma} P^{\text{dis}}_h \big) \Delta t \quad \forall t \in T,  \\
\label{eqn:charging_energy}
    \quad \quad & P_t = P^\text{ch}_t - P^\text{dis}_t \quad \forall t \in T, \\
\label{eqn:charging_bin}
    \quad \quad & P^\text{ch}_t \leq M P^\text{b}_t \quad \forall t \in T,  \\
\label{eqn:discharging_bin}
    \quad \quad & P^\text{dis}_t \leq M (1-P^\text{b}_t) \quad \forall t \in T, \\
\label{eqn:soc_boundary}
    \quad \quad & E^\text{min} \leq E_t \leq E^\text{max} \quad \forall t \in T, \\ 
\label{eqn:charging_limit}
    \quad \quad & -P^\text{max} \leq P_t \leq P^\text{max} \quad \forall t \in T, \\
\label{eqn:daily_cycle}
    \quad \quad & \sum_{t \in T} P^\text{ch}_t \leq E^\text{max}, \\
\label{eqn:net_demand}
    \quad \quad & d_t - g_t + P_t \Delta t = n^+_t - n^-_t \quad \forall t \in T, \\
\label{eqn:import_bin}
    \quad \quad & n^+_t \leq M n^\text{b}_t \quad \forall t \in T, \\
\label{eqn:export_bin}
    \quad \quad & n^-_t \leq M (1-n^\text{b}_t) \quad \forall t \in T, \\
\label{eqn:opt_sign}
    \quad \quad & E_t, P_t, P^\text{ch}_t, P^\text{dis}_t, n^+_t, n^-_t \geq 0 \quad \forall t \in T, \\
\label{eqn:opt_type}
    \quad \quad & P^\text{b}_t, n^\text{b}_t \in \{0,1\} \quad \forall t \in T, &&
\end{flalign}
\end{subequations}
where $\theta = \{E_t, P_t, P^\text{ch}_t, P^\text{dis}_t, n^+_t, n^-_t, P^\text{b}_t, n^\text{b}_t\}$, and $T$ represents the set of time intervals for one day ahead. The first term in \eqref{eqn:obj_func} represents the energy payment at wholesale prices, $\delta_t$, for the net energy exchanged with the grid. If the net energy, $n^+_t - n^-_t$, is positive, the prosumer buys from the grid. Conversely, if the net energy is negative, the prosumer gets paid at wholesale prices for selling their electricity. In addition to paying for electricity at wholesale prices, prosumers are responsible for paying network usage charges, $\lambda_t$, as shown in the second term in \eqref{eqn:obj_func}, which is only applied on the imported energy \citep{ausgridnetworkcharges, amber}. Constraint \eqref{eqn:battery_soc} represents the energy stored, $E_t$, in the battery over the course of one day. Similar to the work in \citep{Optimal2022Dinh}, we consider a round-trip efficiency, $\Gamma$, on the battery (dis)charging cycles. We define the charging, $P^{\text{ch}}_t$, and discharging, $P^{\text{dis}}_t$, power in \eqref{eqn:charging_energy}. To avoid simultaneous charging and discharging, we consider complementarity constraints utilising a binary variable, $P^\text{b}_t$, and a sufficiently large constant, $M$, in \eqref{eqn:charging_bin} and \eqref{eqn:discharging_bin}. Constraint \eqref{eqn:soc_boundary} limits the stored energy within the battery's nominal capacity. Constraint \eqref{eqn:charging_limit} restricts the maximum (dis)charging power of the battery to be lower than its rated power. To mitigate battery degradation caused by excessive (dis)charging, constraint \eqref{eqn:daily_cycle} limits the battery to a maximum of one cycle per day. Constraint \eqref{eqn:net_demand} separates the household consumption, $d_t$, solar PV generation, $g_t$, and the (dis)charging energy of the home battery, $P_t \Delta t$, into imported, $n^+_t$, and exported $n^-_t$, electricity. Similar to the battery (dis)charging power, we restrict $n^+_t$ and $n^-_t$ from simultaneously taking non-zero values using complementarity constraints in \eqref{eqn:import_bin} and \eqref{eqn:export_bin}. Lastly, we define the sign and type of the variables in \eqref{eqn:opt_sign} and \eqref{eqn:opt_type}.

In this optimisation model, wholesale prices are the main source of uncertainty, which can be obtained using the proposed probabilistic forecasts. While there are other sources of uncertainty, such as household consumption, we assume that other uncertain parameters are perfectly known for an easy evaluation of the impact of price uncertainty. To this end, we convert the objective function in \eqref{eqn:obj_func} to an expected cost minimisation problem using expected prices \citep{economic2010zareipour, Julian2021probabilistic}, as follows:
\begin{align}
\label{eqn:obj_func_sce}
    \min_{\theta} E[\text{Cost}] = \sum_{t \in T} E[\delta_{t}] (n^+_t - n^-_t) + \lambda_t n^+_t. &&
\end{align}
If the optimisation is solved using point forecasts, as in AEMO pre-dispatch prices, we can directly substitute the expected prices with these point forecasts. However, to obtain the expected prices from the probabilistic forecasts, we leverage the predictive CDF, outlined in subsection \ref{sec:evaluation_method}, to generate a set of equiprobable samples. We then calculate the expected price as follows:
\begin{align}
\label{eqn:expected_price}
    E[\delta_{t}] = \sum_{s \in S} (\pi_s \hat{\delta}_{t,s}), &&
\end{align}
where $S$ represents the set of price forecast samples, and $\pi_s$ denotes the probability of sample $s \in S$. The solutions obtained from the proposed MILP problem are the optimal scheduled power of the battery systems and the net energy exchanged with the grid for each interval. We calculate the ground truth electricity cost of each household as follows:
\begin{align}
\label{eqn:ground_truth_cost}
    \text{Cost*} = \sum_{t \in T} \delta_{t} (n^{*+}_t - n^{*-}_t) + \lambda_t n^{*+}_t, &&
\end{align}
where $n^{*+}_t, n^{*-}_t$ represent the realised imported energy and realised exported energy, respectively. We use the ground truth results for comparison. 

\begin{table}[!t]
    \caption{Optimisation parameters}
    \centering
    \small
    \begin{tabular}{l|l|l}
    \toprule
    \rowcolor{GrayMid} $\Gamma$ & 90 & Battery round-trip efficiency (\%) \\
    $E^\text{min}$ & 0 & Battery minimum energy (kWh) \\
    \rowcolor{GrayMid} $E^\text{max}$ & 13.5 & Battery maximum energy (kWh) \\
    $E^\text{init}$ & 0 & Battery initial energy (kWh) \\
    \rowcolor{GrayMid} $P^\text{max}$ & 5 & Battery power rating (kW) \\
    $\Delta t$ & 0.5 & Time interval resolution (h) \\
    \rowcolor{GrayMid} $\lambda_t$ & 0.097 & Network usage charge (AUD\$/kWh) \\
    \bottomrule
    \end{tabular}
    \label{tab:simu_input}
\end{table}

\subsection{Optimisation setup}
\begin{itemize}
  \item \textbf{Energy profiles} of ten prosumers were obtained from the Solar Home dataset containing half-hourly solar PV generation and electricity consumption data \citep{Ausgrid2012data}. Since the data resolution is 30 minutes, we consider the price forecasts at the last trading interval of each 30-minute period, similar to subsection \ref{sec:AEMO_price_forecast}.
  \item \textbf{Optimisation period} is the first week of each month in 2021 (12 weeks in total), in which we obtain energy profiles of the prosumers, wholesale electricity prices, and network usage charges \citep{ausgridnetworkcharges}.
  \item \textbf{Battery data} was sourced from the Tesla Powerwall user manual\footnote{Tesla Powerwall 2 Datasheet: \url{https://www.tesla.com/sites/default/files/pdfs/powerwall/Powerwall\%202_AC_Datasheet_en_AU.pdf}}. We summarised the battery and other optimisation parameters in Table \ref{tab:simu_input}.
\end{itemize}

\subsection{Optimisation results}
To evaluate the economic benefit of the proposed Q-QRA model, we compare the ground truth electricity cost of all prosumers using five different sets of wholesale price forecasts, namely, the ground truth wholesale prices (perfect prediction), AEMO pre-dispatch prices, the median forecasts of the Q-QRA model, \textcolor{AdBlue}{quantile} forecasts from QRA model and \textcolor{AdBlue}{quantile} forecasts from Q-QRA model. For each set of probabilistic price forecasts, we use the predictive CDF to generate 100 equally probable expected price samples. Hence, the probability of each sample in \eqref{eqn:expected_price} is set to $\pi_s=0.01$.

\begin{table}[t]
    \caption{The comparison of weekly average electricity costs across 10 prosumers. Negative values indicate profit, with the Ground truth indicating the theoretical upper bound achievable with perfect prediction. Boldface values indicate the most profitable prediction scheme for each Prosumer.}
    \centering
    \begin{tabular}{c|ccccc}
    \toprule
    \rowcolor{GrayTop}\textbf{Prosumer \#} &
    \textbf{Ground truth}&
    \textbf{AEMO pre-dispatch} & \textbf{Q-QRA median}  & \textbf{QRA} & \textbf{Q-QRA} \\
    \midrule
    1 & -6.31 & -0.16 & -1.45 & \textbf{-1.55} & -1.45 \\
    \rowcolor{GrayMid} 2 & 8.61 & 14.90 & 13.92 & 13.90 & \textbf{13.82} \\
    3 & -1.81 & 4.63 & 3.19 & 3.11 & \textbf{3.06} \\
    \rowcolor{GrayMid} 4 & -6.51 & 0.04 & -0.95 & -1.02 & \textbf{-1.12} \\
    5 & -4.43 & 2.06 & 0.95 & 0.89 & \textbf{0.83} \\
    \rowcolor{GrayMid} 6 & 0.27 & 6.69 & 5.30 & 5.19 & \textbf{5.18} \\
    7 & -5.14 & 1.38 & 0.12 & 0.04 & \textbf{-0.03} \\
    \rowcolor{GrayMid} 8 & -3.14 & 3.21 & 2.21 & 2.13 & \textbf{2.05} \\
    9 & -1.62 & 4.61 & 3.74 & 3.69 & \textbf{3.64} \\
    \rowcolor{GrayMid} 10 & -2.73 & 3.70 & 2.83 & 2.70 & \textbf{2.66} \\
    \bottomrule
    \end{tabular}
    \label{tab:electricity_cost}
\end{table}

Table \ref{tab:electricity_cost} compares the weekly average electricity cost of ten prosumers in different price forecast models. Since the model using ground truth wholesale prices would always produce the lowest electricity cost, we highlighted the second lowest electricity cost for easy comparison. The negative cost values indicate the profit of prosumers under the proposed retailer scheme. This profit is achieved mainly by using the home battery to store electricity when prices are low and to export electricity back to the grid when prices are high. Note that prosumers receive payments from the retailer for exporting excess electricity to the grid \citep{amber}. The results in Table \ref{tab:electricity_cost} show that the last three forecast models outperform the AEMO model, with the proposed Q-QRA model providing the least cost solutions for most prosumers. Overall, the results demonstrate that using the price forecasts from the proposed Q-QRA model leads to a minimum cost for most prosumers due to the combined benefits of higher prediction accuracy and consideration of a large range of potential scenarios.

\section{Conclusion}
\label{sec:conclusion}

In this paper, we demonstrated the use of ensemble forecasting methods to predict 24-hour ahead electricity price distribution in the SA region of the Australian NEM. We observed that even when predictions are made using traditionally ‘robust’ models, such as quantile regression, higher accuracy in the inner quantiles can be gained by filtering extreme spikes occurring in the data set. However, in cases where the extreme outer quantiles are concerned, training on the unfiltered series would be more suitable. Next, we demonstrated two post-processing steps, smoothing and autoregression, that can be applied in a probabilistic setting to attain higher predictive accuracy. We showed that probabilistic \textcolor{AdBlue}{forecasting models} could be combined across model classes and training data subsets to create a more accurate prediction process. In terms of the ensemble method, the results from the proposed Q-QRA model indicated a systematic improvement at all quantile levels compared to the constituent models. We further evaluated the superiority of the Q-QRA model in a battery scheduling optimisation problem using real-world data from ten prosumers. The results showed that using Q-QRA forecasts led to the lowest electricity costs for most prosumers.

\section*{Acknowledgements}
This research was supported by the Adelaide Summer Research Scholarship (ASRS), Faculty of Engineering, Computer \& Mathematical Sciences, University of Adelaide.

\section*{Declaration of Generative AI and AI-assisted technologies in the writing process}
During the preparation of this work, the authors used ChatGPT to improve readability and language. After using this tool/service, the authors reviewed and edited the content as needed and take full responsibility for the content of the publication.

\bibliography{camsbib}

\end{document}